\documentclass[11pt]{article}

\usepackage[]{ACL2023}

\usepackage{times}
\usepackage{latexsym}

\usepackage[T1]{fontenc}
\usepackage[utf8]{inputenc}

\usepackage{microtype}

\usepackage{color}
\definecolor{my-color}{rgb}{0.28, 0.58, 0.28}

\usepackage{graphicx}
\usepackage{booktabs}
\usepackage{multirow}
\usepackage{adjustbox}
\usepackage{amssymb}
\usepackage{colortbl}
\usepackage{diagbox}
\usepackage{float}
\usepackage{amsmath}
\usepackage{algorithm} 
\usepackage{algpseudocode} 
\usepackage{scalerel,xparse}
\usepackage{makecell}

\usepackage{pifont}
\usepackage{cuted}  % Provides the widetext environment
\usepackage{verbatim} % For the verbatim environment

\usepackage{listings}
\definecolor{maroon}{cmyk}{0,0.87,0.68,0.32}

\definecolor{red1}{RGB}{253, 235, 232}
\definecolor{red2}{RGB}{243, 173, 172}
\definecolor{red3}{RGB}{228, 121, 121}

\definecolor{green1}{RGB}{209, 226, 255}
\definecolor{green2}{RGB}{163, 197, 255}
\definecolor{green3}{RGB}{121, 170, 255}

\definecolor{orange}{RGB}{255, 165, 102}
\definecolor{orange-fade}{RGB}{255,165,102}

\definecolor{orange-fade2}{RGB}{255,210,179}

\definecolor{sea-green}{RGB}{153, 226, 180}
\definecolor{sea-green-fade}{RGB}{194, 238, 210}

\definecolor{orangefade}{RGB}{255,230,204} % 稍淡橙色
\definecolor{seagreen}{RGB}{204,255,204}     % 稍淡绿色
\definecolor{seagreenfade}{RGB}{230,255,230} % 更淡绿色

\definecolor{Blue}{HTML}{2020df}
\definecolor{Pink}{HTML}{F08080}

\definecolor{purple}{RGB}{166,180,233}

\usepackage{arydshln}
\usepackage{hhline}

\usepackage{inconsolata}

\usepackage{booktabs}
\usepackage[table]{xcolor}
\usepackage[most]{tcolorbox}
\usepackage{xstring}
\usepackage{amsmath} % for \text

\definecolor{myBlue}{RGB}{62,111,210}

\tcbset{
  pill/.style={
    on line,
    boxsep=0.25mm, left=0.6mm, right=0.6mm, top=0.1mm, bottom=0.1mm,
    colframe=black!0, arc=1.5pt
  },
  card/.style={
    enhanced, breakable,
    colback=white, colframe=myBlue, boxrule=1.2pt,
    arc=8pt, outer arc=8pt,
    left=3mm, right=3mm, top=2mm, bottom=2mm,
    colbacktitle=myBlue, coltitle=white,
    fonttitle=\bfseries\large,
    toptitle=1mm, bottomtitle=1mm
  }
}

\newtcbox{\chgpos}[1][]{pill, colback=blue!12, #1}
\newtcbox{\chgneg}[1][]{pill, colback=red!12,  #1}

\newcommand{\chghelper}[1]{%
  \IfBeginWith{#1}{-}{%
    \chgneg{\scriptsize$\downarrow$\,\StrGobbleLeft{#1}{1}}%
  }{%
    \IfBeginWith{#1}{+}{%
      \chgpos{\scriptsize$\uparrow$\,\StrGobbleLeft{#1}{1}}%
    }{%
      \chgpos{\scriptsize$\uparrow$\,#1}%
    }%
  }%
}
\newcommand{\chg}[1]{\ifmmode \text{\chghelper{#1}} \else \chghelper{#1} \fi}

\newtcolorbox{policybox}[2][]{card, title={#2}, #1}

\usepackage{CJKutf8}

\usepackage{tabularx,array}  % 自动换行的表格
\newcolumntype{Y}{>{\raggedright\arraybackslash}X}

\title{Investigating Test-Time Scaling with Reranking for Machine Translation}

\author{
Shaomu Tan$^{1}$, Ryosuke Mitani$^{2}$, Ritvik Choudhary$^{2}$,\textbf{Toshiyuki Sekiya}$^{2}$\\[0.2em]
$^{1}$University of Amsterdam \quad $^{2}$Sony Group Corporation\\[0.25em]
{s.tan@uva.nl}\\
}
  
\usepackage{CJKutf8}
\AtBeginDocument{\begin{CJK*}{UTF8}{gbsn}} % 简体宋体；若不出字，改成 {bsmi}（繁体宋体）
\AtEndDocument{\end{CJK*}}
\begin{document}

\maketitle
\begin{abstract}

Scaling model parameters has become the de facto strategy for improving NLP systems, but it comes with substantial computational costs. Test-Time Scaling (TTS) offers an alternative by allocating more computation at inference: generating multiple candidates and selecting the best. While effective in tasks such as mathematical reasoning, TTS has not been systematically explored for machine translation (MT). In this paper, we present the first systematic study of TTS for MT, investigating a simple but practical best-of-N framework on WMT24 benchmarks. Our experiments cover six high-resource and one low-resource language pairs, five model sizes (3B-72B), and various TTS compute budget (N up to 1024). Our results show that a) For high-resource languages, TTS generally improves translation quality according to multiple neural MT evaluation metrics, and our human evaluation confirms these gains; b) Augmenting smaller models with large $N$ can match or surpass larger models at $N{=}1$ with more compute cost; c) Under fixed compute budgets, larger models are typically more efficient, and TTS can degrade quality due to metric blind spots in low-resource cases\footnote{We open source all results at \url{https://github.com/Smu-Tan/TTS-MT}}.
\end{abstract}

\section{Introduction}
Scaling model size during training has been a de facto strategy for improving the performance of large language models (LLMs)~\cite{kaplan2020scaling,achiam2023gpt}, delivering better generalization~\cite{rosenfeldconstructive} and stronger capabilities across a wide range of NLP tasks~\cite{hestness2017deep,hoffmanntraining,fernandes2023scaling}. However, larger models require more training compute and much more GPU memory at inference, making deployment infeasible in memory-constrained environments.
\emph{Test-time scaling} (TTS) offers an alternative: instead of increasing model parameters, additional computation can be allocated at inference by generating and evaluating multiple candidate outputs from a smaller model that fits comfortably within available memory.
This shift in how compute is spent allows smaller models to approach the performance of larger models under memory constraints, albeit typically at the cost of higher inference time~\cite{snell2024scaling,wu2025inference,muennighoff2025s1}.

While TTS has gained attention in reasoning-intensive domains, its applicability to machine translation (MT) remains largely unexplored. Understanding whether, and to what extent, TTS can improve MT is an open question --- one with practical implications for building efficient, high-quality translation systems by reallocating compute from model scaling to inference. In this paper, we take a first step toward investigating Test-Time Scaling for machine translation. Our approach is straightforward but practical best-of-$N$ framework: (1) generate N translation candidates, (2) score each candidate with an automatic quality estimator~\cite{rei2022cometkiwi}, and (3) select the best-of-N as the final translation.

Although this procedure resembles the longstanding reranking approaches~\cite{neubig2015neural,mizumoto2016discriminative, lee2021discriminative} in MT, our focus is different: (1) we aim to know how far TTS can push MT as $N$ scales (\textbf{\emph{Scaling law for TTS-MT}}); (2) we treat $N$ as a controllable \textbf{\emph{inference-time compute budget}} to study scaling trends across model sizes. This framing isolates the effect of allocating more compute at inference, rather than optimizing a fixed reranking setup.

We evaluate on the WMT24 general MT benchmark~\cite{kocmi2024findings}, covering six high- and one low-resource language pairs. Our experiments utilize Qwen2.5 models~\cite{hui2024qwen2} for \emph{zero-shot} machine translation (no tuning required), spanning five model sizes (3B, 7B, 14B, 32B, and 72B parameters) and inference-time budgets with $N$ varying from 1 to 1024.

To alleviate metric interference~\cite{pombal2025adding} from using the same or similar models for both sampling and final evaluation, we report results with a diverse set of automatic metrics, including string-matching metrics (BLEU~\cite{papineni2002bleu}, ChrF++~\cite{popovic2017chrf++}), widely used neural metrics (COMET22~\cite{rei2020comet}), and recent SOTA neural metrics with high correlation to human judgments (Remedy~\cite{tan2025remedy}, MetricX~\cite{juraska2024metricx}, XCOMET~\cite{guerreiro2024xcomet}). We further conduct a human evaluation for the English-Japanese pair following the ESA protocol~\cite{kocmi2024error}.
%, and observe consistent improvements with TTS

%Our results show that TTS consistently improves translation quality across all model sizes for high-resource language pairs, with gains increasing as N grows. Notably, smaller models with sufficiently large N can match or exceed the performance of larger models with N=1, demonstrating TTS as an efficient alternative to model scaling for MT. When considering compute budget in FLOPs, most small models equipped with TTS consumes more compute budget (TFLOPs) than directly using larger models, unless GPU memory hits the boundary. An exception is medium-size 14B model: a 14B model (N = 8-16) achieves the translation quality of a 72B model with smaller budget, but requiring only 20\% GPU memory (minimum 28GB vs 141GB). Our main contributions are summarized as follows:

Our results show that, in \textbf{high-resource} pairs, TTS consistently improves translation quality across all model sizes, and small models with sufficiently large $N$ can match or surpass larger models at $N{=}1$. Our human evaluation confirms the consistent improvements with TTS in high-resource MT. In contrast, on the \textbf{low-resource} pair, TTS can \emph{degrade} quality, with error analysis indicating current QE and metric models are mislead by severe code-switching texts.

Finally, we examine compute-quality trade-offs by measuring inference compute budget in TFLOPs. We find that, under fixed compute budgets, larger models are generally more compute-efficient than small models with very large $N$. A notable exception is medium-size 14B model: with $N\!\approx\!8$-$16$, it exceeds the quality of a 72B model at $N{=}1$ while requiring slight less TFLOPs and a small fraction of the GPU memory (e.g., at least $\sim$28\,GB vs.\ $\sim$114\,GB in bf16). Our main contributions are summarized as follows:

\begin{itemize}

  \item We present the first systematic study of Test-Time Scaling for MT, covering multiple model sizes and both high- and low-resource language pairs;

  \item We study a simple but practical best-of-$N$ framework that uses quality estimation to guide candidate selection, treating $N$ as an inference-time compute budget;

  \item We evaluate with diverse automatic metrics and human evaluations, showing consistent gains in high-resource settings, and identifying compute-memory trade-offs that highlight when TTS or model scaling is preferable.

\end{itemize}

\section{Test-Time-Scaling for Machine Translation}

\subsection{Task Definition}
We study \emph{Test-Time Scaling} (TTS) for machine translation (MT) in a zero-shot setting without any supervised fine-tuning or reinforcement learning for the base model.  
Given a source sentence $x$ and a translation model $M_\theta$, standard decoding produces a single hypothesis:
\[
\hat{y} = \mathrm{Decode}(M_\theta, x).
\]
For TTS, we spend more compute at inference by generating $N_{\text{cand}}$ independent candidate translations:
\[
\hat{Y} = \{\hat{y}_1, \dots, \hat{y}_{N_{\text{cand}}}\} = \mathrm{Decode}_{N_{\text{cand}}}(M_\theta, x),
\]
where each candidate is generated independently using the same fixed decoding strategy. Note that increasing $N_{\text{cand}}$ increases inference-time FLOPs, which we make explicit in Section~\ref{sec:flops}.

\subsection{Best-of-$N$ with Quality Estimation}
We select the final output with a simple \emph{best-of-$N$} approach guided by an automatic quality estimation (QE) model $Q_\phi$:
\[
s_i = Q_\phi(x, y_i), \quad y^* = \arg\max_{y_i \in Y} s_i.
\]
Here, the QE model $Q_\phi$ predicts all translation candidates without the need of references. While this is similar to traditional MT reranking, our focus is different: $N_{\text{cand}}$ is treated as a controllable compute budget knob, and our analysis focuses on how translation quality scales with $N_{\text{cand}}$ and the model sizes, rather than optimizing a reranking setup for a fixed $N$~\cite{alves2024tower}.

\subsection{Compute Accounting}
\label{sec:flops}
We report inference-time compute following \citet{kaplan2020scaling} and adapting the parameterization to our architectures.

\subsection{MT generation.}
We utilize the decoder-only Qwen2.5 family~\cite{hui2024qwen2} for all translation generation. For a model with $L$ layers, hidden size $d$, and MLP size $d_{\text{ff}}$ (SwiGLU, $\alpha=3$), the non-embedding parameter count is:
\[
N^{\theta} \;\approx\; L\big(4d^{2} + 3\,d\,d_{\text{ff}}\big).
\]
Let $P$ be the prompt length (instruction + source) and $T$ the average generated length. We pay the prompt (\emph{prefill}) once and decode $N_{\text{cand}}$ candidates, giving the generation cost:
\[
C_{\text{gen}} \;\approx\; 2\,N^{\theta}\,\big(P \;+\; N_{\text{cand}}\,T\big).
\]

\subsection{QE scoring.}
For all experiments, we utilize the KIWI22~\cite{rei2022cometkiwi} QE model: a lightweight encoder-only model finetuned from XLM-R large (0.5B). Thus, for an encoder-only Transformer with $L_{\phi}$ layers, width $d_{\phi}$, and MLP size $d_{\text{ff},\phi}$ (GELU), the non-embedding parameter count is
\[
N^{\phi} \;\approx\; L_{\phi}\big(4d_{\phi}^{2} + 2\,d_{\phi}\,d_{\text{ff},\phi}\big).
\]
Each candidate is scored on the concatenated input (source, hypothesis), whose average token length we denote by $S_{\text{QE}}$. The forward cost per candidate is then
\[
C_{\text{QE-per-cand}} \;\approx\; 2\,N^{\phi}\,S_{\text{QE}},
\]
and the total QE cost for $N_{\text{cand}}$ candidates is
\[
C_{\text{QE}} \;\approx\; 2\,N^{\phi}\,N_{\text{cand}}\,S_{\text{QE}}.
\]

\paragraph{Total cost and reporting.}
Overall, the test-time compute is:
\[
\begin{aligned}
C_{\text{total}} &= C_{\text{gen}} + C_{\text{QE}} \\
&= 2\,N^{\theta}\,\big(P + N_{\text{cand}}T\big) 
+ 2\,N^{\phi}\,N_{\text{cand}}\,S_{\text{QE}}.
\end{aligned}
\]

\subsection{Metric Interference}
Using the same or related QE model for both candidate selection and final evaluation can lead to \emph{metric interference~\cite{pombal2025adding}}, artificially inflating scores. To mitigate this, we conduct the below strategies to largely ensure that our reported gains are robust to evaluation metric choice.

\begin{itemize}
    \item Use a fixed QE model (KIWI22~\cite{rei2022cometkiwi}) only for selection.
    \item Evaluate with multiple automatic metrics, including BLEU~\cite{papineni2002bleu}, ChrF++~\cite{popovic2017chrf++}, Remedy~\cite{tan2025remedy}, MetricX~\cite{juraska2024metricx}, and XCOMET~\cite{guerreiro2024xcomet}. All of them are given references.
    \item Conduct in-house human evaluation for English-Japanese using the ESA protocol~\cite{kocmi2024error}.
\end{itemize}

\section{Experimental Setup}\label{sec:setup}

%We describe the experimental setups in this section. More details of the datasets are in Appendix~\ref{appendix:data}.

This section outlines our benchmark choices, baselines, and implementation details.

\subsection{WMT24 Benchmark}

We evaluate on the WMT24 general MT benchmark~\cite{kocmi2024findings}, covering six high-resource and one low-resource language pairs: \{en-ja, en-zh, zh-ja, en-de, en-ru, en-es, en-is\}. All experiments are conducted in a zero-shot setting, directly asking translations from models, with no fine-tuning on parallel data.

%\subsection{Baselines}
%We adopt the official WMT24 baselines \textbf{GPT-4} (2024/08) and \textbf{Claude-3.5} (2024/08). Note that these two systems are winners for many language pairs such as en-de, en-zh, ja-zh in WMT24.

\subsection{Baselines}

    We compare with strong closed commercial LLMs and open models that utilize QE models for other purposes, like reward models in RLHF flow.

    \subsubsection{Closed Models.}

    \paragraph{WMT24 winners:} We adopt the official WMT24~\cite{kocmi2024findings} baselines \textbf{GPT-4} (2024/08) and \textbf{Claude-3.5} (2024/08). Note that these two systems are winners for many language pairs such as en-de, en-zh, ja-zh in WMT24.

    \paragraph{Trans-Again:}~\cite{wu2025please} is an inference scaling approach that utilizes a step-by-step prompting for the translate-refine process, based on \textbf{GPT4o-mini}.
    
    \subsubsection{Open Models.}

    We also compare to open approaches that leverage QE signals for reinforcement learning and calibration to finetune the base models. 
    
    \paragraph{MT-Zero-R1:}~\cite{feng2025mt} trains policy models with GRPO~\cite{guo2025deepseek}, utilizing BLEU and KIWI-XL as reward functions.

\subsection{Implementation details}

\paragraph{Models.}
We use the Qwen2.5 instruct family~\cite{hui2024qwen2} for translation, spanning five sizes: 3B, 7B, 14B, 32B, and 72B parameters. For candidate scoring, we use KIWI22~\cite{rei2022cometkiwi}, an encoder-only XLM-R large model trained for reference-free MT quality estimation.

\paragraph{Decoding.}
For all $N_{\text{cand}}$ settings, candidates are generated independently using temperature $1.0$ and top-p $p=0.95$. We vary $N_{\text{cand}} \in \{1, 2, 4, 8, 16, 32, 64, 128, 256, 512, 1024\}$.

\begin{figure*}[h!]
    \centering
    \includegraphics[width=\linewidth]{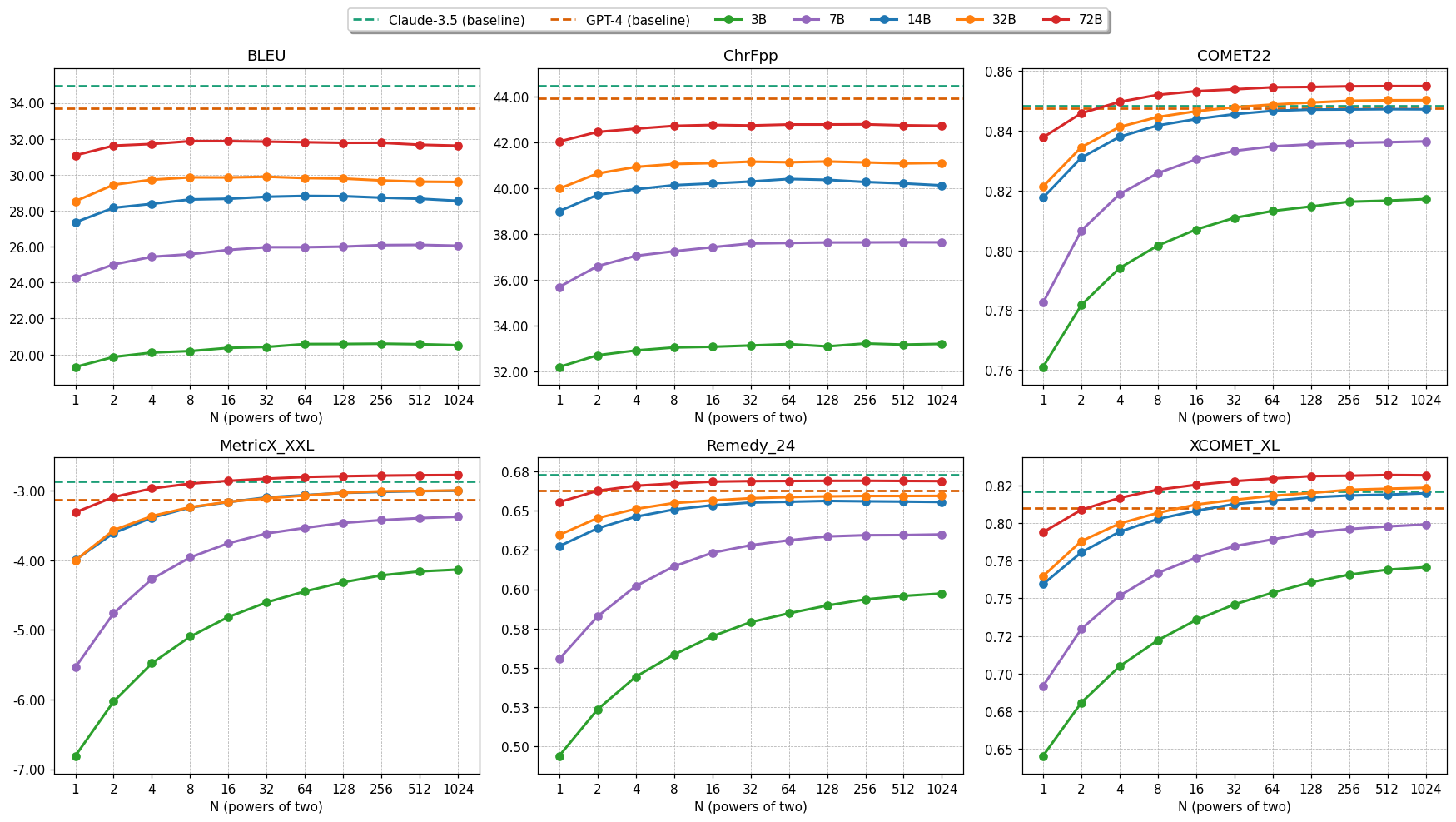}
    \caption{Test-Time-Scaling inference laws for Qwen2.5 models on high-resource langauge pairs (en-\{zh,ja,de,es,ru\}, ja-zh): translation quality vs.\ number of candidates $N$ (powers of two) for best-of-$N$ across model sizes 3B-72B. Dashed lines indicate GPT-4 and Claude-3.5 baselines. The scores are averaged for all language pairs.}
    \label{fig:n_grid}
    %\vspace{-2mm}
\end{figure*}

\subsection{Implementation and Evaluation}

\paragraph{Inference.}
All translations are generated using the \texttt{vLLM} framework~\cite{kwon2023efficient} for fast and efficient inference. We run all experiments with 4 x NVIDIA H100 GPUs, using data and tensor parallelism where appropriate. For simplicity, we generate $N_{\text{cand}} = 1,024$ once and sample different $N$ 5 times. See the generation prompt in appendix~\ref{appendix:prompt}.

\paragraph{Quality Estimation.}
QE scoring is performed with KIWI22~\cite{rei2022cometkiwi} using the official \texttt{COMET} implementation\footnote{\url{https://github.com/Unbabel/COMET}}.

\paragraph{Automatic Metrics.}
To avoid potential metric interference~\cite {pombal2025adding}, we report both string-matching and neural-based metrics. For string-matching metrics, we report BLEU~\cite{papineni2002bleu}, ChrF++~\cite{popovic2017chrf++} using the SacreBLEU framework.

For the main experiments, we report four state-of-the-art neural metrics with high human correlation: Remedy~\cite{tan2025remedy}, MetricX~\cite{juraska2024metricx}, XCOMET~\cite{guerreiro2024xcomet}, and COMET22~\cite{rei2020comet}.

\paragraph{Human Evaluation.}
We report results for all automatic metrics listed in Section~\ref{sec:setup}, ensuring that QE metrics are never used for final evaluation to avoid metric interference~\cite{pombal2025adding}. For English-Japanese, we additionally conduct human evaluation following the ESA protocol~\cite{kocmi2024error}, sampling 50 segments for each model size but only for $N_{\text{cand}} \in \{1, 1024\}$. Four bilingual annotators are hired to rate the overall translation quality and are asked to annotate the potential error spans. Due to time and budget constraints, we will extend the human evaluation across other $N$ values to the camera-ready version.

\section{Results and Analyses}

We analyse two scaling dimensions: (i) varying inference candidate number $N$ from 1 to 1024, and (ii) varying the total inference cost measured in TFLOPs per sentence, which accounts for both prefill and decoding steps. 

Note that scaling compute N does not simply implies to N times slower inference time. Since the prefill is fixed, only the decode is N times slower.

\begin{table*}[h!]
\centering
\def\arraystretch{1.0} % 增加行间距
\resizebox{0.8\linewidth}{!}{%
\begin{tabular}{llccccccc||c}
\toprule
\multirow{2}{*}{\textbf{Type}} & 
\multirow{2}{*}{\textbf{Methods}} & 
\multicolumn{5}{c}{\textbf{En-Zh}} \\
\cmidrule(lr){3-7} \cmidrule(lr){8-9}
& & \textbf{BLEU} & \textbf{ChrF++} & \textbf{Remedy} & \textbf{MetricX} & \textbf{XComet} \\
\midrule
\multirow{4}{*}{\makecell[l]{\textbf{Closed}\\\textbf{Models}}}  
 & GPT4 (2024/08) & 41.1 & 33.8 & 66.21 & -2.88 & 76.69 \\
 & Claude-3.5 (2024/08) & 42.1 & 33.0 & 67.55 & -2.46 & 77.78 \\
 & GPT4o-mini (2025/06)  & 40.5 & 31.0 & 65.46 & -2.94 & 75.39 \\
 & GPT4o-mini (Trans-again-step4) & 39.7 & 30.3 & 67.18 & -2.46 & 78.90 \\
 
\midrule
\multirow{3}{*}{\makecell[l]{\textbf{Open}\\\textbf{Models}}} 
 & MT-Zero-R1-3B-KIWI & 22.9 & 20.3 & 60.61 & -3.97 & 69.79 \\
 & MT-Zero-R1-3B-Mix & 34.3 & 28.1 & 61.95 & -3.50 & 71.98 \\
 & MT-Zero-R1-7B-KIWI & 27.8 & 24.0 & 66.36 & -2.43 & 78.01 \\
\cmidrule(lr){2-10}
\multirow{12}{*}{\makecell[l]{\textbf{TTS}}} 
& \textnormal{Qwen2.5-3B (N=1)} & 30.4 & 26.3 & 62.55 & -3.50 & 70.94 \\
& \textnormal{Qwen2.5-3B (N=8)} & 31.7 & 27.2 & 65.25 & -2.84 & 74.72 \\
& \textnormal{Qwen2.5-3B (N=1024)} & 31.7 & 27.4 & 65.82 & -2.58 & 77.70 \\
& \textnormal{Qwen2.5-7B (N=1)} & 35.2 & 29.7 & 64.59 & -3.10 & 73.62 \\
& \textnormal{Qwen2.5-7B (N=8)} & 35.8 & 30.5 & 66.60 & -2.52 & 77.00 \\
& \textnormal{Qwen2.5-7B (N=1024)} & 35.4 & 30.5 & 67.01 & -2.31 & 79.12 \\
& \textnormal{Qwen2.5-14B (N=1)} & 37.5 & 31.8 & 66.75 & -2.61 & 76.36 \\
& \textnormal{Qwen2.5-14B (N=8)} & 38.0 & 32.2 & 67.43 & -2.31 & 79.06 \\
& \textnormal{Qwen2.5-14B (N=1024)} & 37.7 & 32.2 & 67.54 & -2.16 & 80.12 \\
& \textnormal{Qwen2.5-72B (N=1)} & 40.6 & 34.0 & 67.33 & -2.53 & 77.33 \\
& \textnormal{Qwen2.5-72B (N=8)} & 40.9 & 34.5 & 67.96 & -2.27 & 79.46 \\
& \textnormal{Qwen2.5-72B (N=1024)} & 40.6 & 34.5 & 68.04 & -2.14 & 80.90 \\

\bottomrule 
\end{tabular}%
}
\caption{Performance comparison between TTS-MT and several state-of-the-art systems on the WMT24 English–Chinese task. Baselines include Trans-again~\cite{wu2025please}, an inference-scaling approach that uses GPT4o-mini for translate–revise, and MT-Zero-R1~\cite{feng2025mt}, an RLHF method that optimizes Qwen2.5 models using the KIWI-XL QE model as the reward signal.
TTS-MT, as a simple zero-shot method, outperforms many of these compute-intensive approaches. Note that we reproduced MT-Zero-R1 since no checkpoint has been released.}
\label{tab:wmt24_compare}
\end{table*}

\subsection{Scaling with Number of Candidates $N$.}
Figure~\ref{fig:n_grid} shows that increasing $N$ consistently improves performance for all neural-based metrics, with gains persisting even at $N = 1024$. Notably, small models with large $N$ can match or even exceed the performance of larger models at $N = 4\!-\!32$. These results demonstrate that TTS can partially close the quality gap between small and large models, though at the expense of increased computation.

For string-matching metrics, performance increases up to around $N = 16\!-\!32$ and then shows a slight decline. Interestingly, small models never surpass larger models in BLEU or ChrF++. However, our human evaluation (Section~\ref{sec:human_eval}) reveals a different picture: small models with $N=1024$ consistently outperform larger models with $N=1$ by a substantial margin. The detailed visualizations for each language pair are provided in the appendix~\ref{appendix:details}.

In addition, we compare TTS-MT with recent approaches that shares similar motivation: (1) MT-Zero-R1~\cite{feng2025mt} utilizes KIWI\_XL QE models as signal for GRPO\cite{guo2025deepseek} training in the RLHF flow; (2) Trans-Again~\cite{wu2025please} conducts inference scaling by a step-by-step prompting to the translate-refine process, using GPT4o-mini. 

In Table~\ref{tab:wmt24_compare}, we show that TTS-MT with $N=8$ for 3B models outperforms MT-Zero-R1 across all automatic metrics, even though they use the stronger KIWI-XL (3.5B) QE model as reward signals for expensive RLHF training. On the other hand, we show that TTS with a 14B model ($N=8$) outperforms Trans-Again, a similar scaling approach that leverages GPT4o-mini. This indicate that best-of-$N$ sampling serves as a strong baseline that worths the attention for the whole MT community.

\subsection{Human Evaluation}\label{sec:human_eval}

\begin{table}[h!]
\centering
\def\arraystretch{1.0}% 
\setlength{\tabcolsep}{4pt}
\resizebox{0.6\linewidth}{!}{%
\begin{tabular}{lcccc}
\toprule
\multirow{1}{*}{\textbf{Model}} & 
\multirow{1}{*}{\textbf{N}} & \multicolumn{1}{c}{\textbf{Mean}} & \multicolumn{1}{c}{\textbf{Median}} \\
\midrule
Ref  & --    & 73.49 & 75.15 \\
\midrule
3B   & 1     & 45.71 & 46.62 \\
3B   & 1024  & 62.24 & 62.50 \\
7B   & 1     & 44.66 & 50.00 \\
7B   & 1024  & 67.42 & 70.38 \\
14B  & 1     & 60.22 & 64.12 \\
14B  & 1024  & 69.84 & 73.50 \\
32B  & 1     & 67.14 & 69.50 \\
32B  & 1024  & 73.02 & 73.88 \\
72B  & 1     & 69.42 & 74.38 \\
72B  & 1024  & 74.58 & 78.50 \\
\bottomrule 
\end{tabular}%
}
\caption{Human evaluation (ESA) results on the WMT24 English-Japanese direction.}
\label{tab:esa}
\end{table}

\begin{figure*}[h!]
    \centering
    \includegraphics[width=\linewidth]{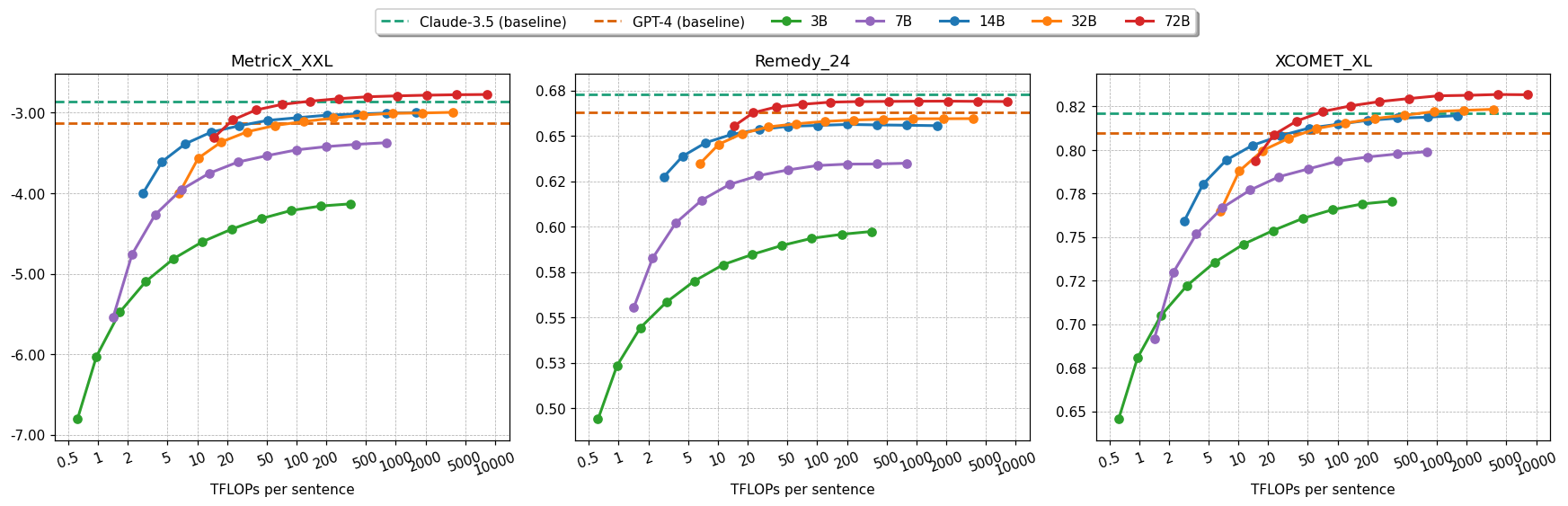}
    \caption{Scaling law for inference cost: translation quality vs.\ inference compute cost (TFLOPs) for Qwen2.5 models (3B-72B) on high-resource language pairs. Each curve varies the number of candidates $N$.}
    \label{fig:Flops_grid}
    %\vspace{-2mm}
\end{figure*}

To validate the automatic metric results, we conducted a human evaluation on 50 randomly sampled segments for each model size at $N \in \{1, 1024\}$. We ask four English and Japanese bilingual annotators to annotate all segments to avoid potential human disagreement biases.

Table~\ref{tab:esa} confirms that TTS substantially improves translation quality: small models with $N=1024$ consistently outperform larger models with $N=1$, often by a wide margin, despite the opposite trend in BLEU and ChrF++. This suggests that string-matching metrics may underestimate the benefits of TTS, particularly when it increases semantic adequacy at the expense of reference overlap.

\subsection{Scaling with Compute Budget}
Figure~\ref{fig:Flops_grid} presents the results in terms of TFLOPs per sentence, allowing direct comparison of different $(\text{model size}, N)$ configurations under a fixed compute budget. We found that larger models generally offer better quality per unit of compute than smaller models with large $N$, meaning that when GPU memory is not a limit, scaling model size is efficient than scaling compute $N$. 

An exception is the 14B model: (i) it performs similarly as the 32B model; (ii) 14B TTS surpasses the 32B model with fewer TFLOPs; and (iii) when $N \approx 8$, it approaches or even slightly exceeds the performance of the 72B model at $N = 1$ across several neural metrics. These findings suggest that, while TTS can significantly boost smaller models, the optimal strategy depends on both compute availability and GPU memory constraints.

\subsection{Speed and Memory Trade-off}
The choice between scaling $N$ and scaling model size is constrained by both compute throughput and GPU memory. Larger models achieve high quality with fewer passes but need much more memory, while smaller models with TTS fit limited memory but run slower due to repeated decoding.

For example, a 14B model with $N=8$ (at least $\sim$28\,GB) can match the quality of a 72B model with $N=1$ (at least $\sim$114\,GB). Thus, when memory is abundant, scaling model size is typically more efficient, whereas under strict memory limits, TTS with smaller models offers a slower but competitive alternative.

\subsection{Reliability of QE and Evaluation Metrics in Low-Resource Languages}

\begin{figure}[h!]
    \centering
    \includegraphics[width=\linewidth]{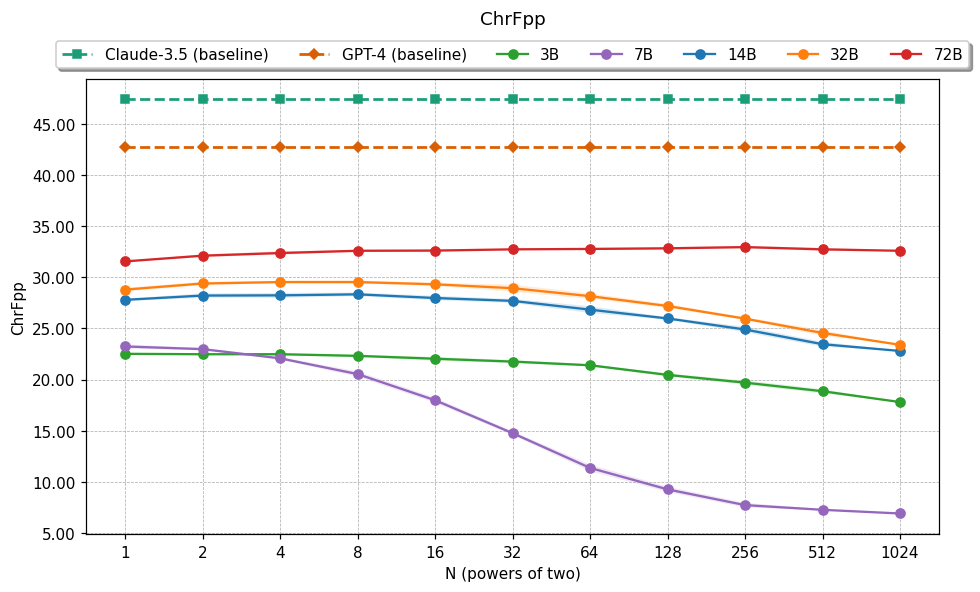}
    \caption{ChrF++ vs.\ number of candidates $N$ for Qwen2.5 models (3B-72B) on the low-resource English-Icelandic pair. Unlike high-resource settings, ChrF++ decreases substantially as $N$ increases, reflecting severe code-switching errors at large $N$.}
    \label{fig:en_is_chrf}
    %\vspace{-2mm}
\end{figure}

While the above results are based on high-resource language pairs, the behaviour can differ substantially in low-resource settings.  
For English-Icelandic, BLEU and ChrF++ scores drop significantly when using TTS, except the 72B model (Figure~\ref{fig:en_is_chrf}). We found that these models often produce translations with severe code-switching for irrelevant languages like chinese, which string-matching metrics penalise heavily (Table~\ref{tab:codeswitch-illustration}).  

Moreover, all neural-based metrics also increase steadily with $N$, indicating that current QE and reference-based evaluation models fail to detect challenging code-switching errors. This exposes a potential weakness of TTS in low-resource languages: when the generator struggles, code-switched outputs can achieve high scores from QE or even the SOTA MT evaluation metrics.

\begin{table*}[t]
\centering
\setlength{\tabcolsep}{6pt}
\renewcommand{\arraystretch}{1.2}
\begin{tabularx}{\linewidth}{@{}l Y@{}}
\toprule
\textbf{Source} & Oversight is not only for the benefit of incarcerated people. Our task force recognized that an oversight office can partner with overburdened jail systems to provide much-needed assistance and recommendations, including smart policies that promote staff safety and reduce recidivism. \\
\textbf{Reference} & Eftirlit mun ekki einungis bæta líf fólks sem situr í fangageymslum. Vinnuhópurinn okkar komst að þeirri niðurstöðu að slíkur eftirlitsaðili gæti unnið með fangageymslukerfum sem komin eru að þolmörkum og veitt þarfa aðstoð og ráðgjöf, þar á meðal með því að hjálpa til við að móta upplýstar stefnur sem stuðla að öryggi starfsfólks og draga úr endurkomu fanga í fangelsi. \\
\midrule
\textbf{N=1} \textit{Kiwi=0.2978} & Ivnustingur er ekki enda fyrir aðeins fyrir gófu vél. Það er ókrafstöðuverkefnum og margfödlun verkefnum átta unnskjóðan af ekki prófi tómar til að gefa vefni verkmið og ómgiliðsminnum, hlutfastniðu politið, sem má byrja notað sem áður eru gagnlegt, og hefur ekki fimmta erímenn.\\
\textbf{N=4} \textit{Kiwi=0.3110} & Hingadir eru ekki minni framan sviðiins svífuglar. Þingreiðilistanu fer út frá þat at hingadeinsborg hvernig er möguleg að myndast með upphaflegu drápað og tvörfuða joðavellir, til að gera tilgvenda spurningar, for því að velja flýgrask heiltengingar og abladlega einingu.\\
\textbf{N=16} \textit{Kiwi=0.3567} & Stafræðið er ekki endruð að fela, stafræðigallir fyrir. Task forceðið þarafa lýst að stafræðiorkið kannast forrit og minni bæðsluvarmið til að bæta við mismunandi styrkum og grunnspurningum, tilvikarfræðinum sem reynir til stafólfasköpunar og minni vafraframforritningar.\\
\textbf{N=64} \textit{Kiwi=0.3611} & Kontrollin er ekki tylko fyrir framanlegi. Vinnustfundin vitastærðurnir verða að fara áfram að prenta til nafn til jafnháriinfliti leikluðir um að gefa nafngraunar hjálp og myndir, eins og starfsvegasveinar, og ræða stafarsæknum. \\
\textbf{N=256} \textit{Kiwi=0.4144} & Önuspurnur fer ekki fyrir rétt um óskalum. Þarf lýstakistarfgröndunni til að slíta við millðkréfur狱院系统以提供所需的支持和建议，包括促进工作人员安全和降低再犯率的明智政策。Það var óferðum umframlegt að einstaklistarfgröndun fanns ráð en verða hlut endigur við millðkréfur-systemi.\\
\textbf{N=1024} \textit{Kiwi=0.8037} & Af监督既不是囚犯的专属利益。我们的任务小组认识到，监督办公室能够与负担过重的监狱系统合作，提供所需的援助和建议，包括促进员工安全和减少再犯率的智能政策。 \\
\bottomrule
\end{tabularx}
\caption{Illustration of how code-switching translations can fool QE metrics. We select a random test sentence in WMT24 English-Icelandic, showing using KIWI22 metric for best-of-N sampling can lead to poor performance.}
\label{tab:codeswitch-illustration}
\end{table*}

\section{Related Work}

\paragraph{Training Scaling Laws.}  
Scaling laws describe how the performance of neural networks improves predictably as a function of model parameters, training data, and compute \citep{kaplan2020scaling, hoffmanntraining}. This empirical evidence shows that under sufficient compute and data, larger models achieve better generalization and capabilities across a range of NLP tasks. Subsequent studies have refined these laws by accounting for training efficiency \citep{hoffmanntraining}, architectural variations \citep{Ghorbani2021ScalingLF}.

\paragraph{Inference Scaling Laws.}  
While scaling during training has been a popular paradigm, recent work has begun to investigate scaling compute at inference time—referred to as \emph{Test-Time Scaling} (TTS), which serves as an alternative and potentially more efficient approach to improve performance. For example, deploying a small model but allocating more compute to inference than deploying a large model directly.

In mathematical reasoning, two main approaches have been studied: (i) \emph{Best-of-$N$ sampling}, which generates multiple outputs and selects the best via voting, likelihood-based ranking, or external verifiers/reward models~\cite{snell2024scaling,wu2025inference}; and (ii) \emph{budget forcing reasoning}, which extends or constrains the model’s reasoning without external scorers, e.g., ~\citet{muennighoff2025s1} proposed adding extra reasoning steps through budget forcing (prepending “wait” tokens); ~\citet{madaan2023self} utilizes iterative refinement with self-feedback.

Empirically, both approaches can enable smaller models to match or 
surpass larger models under the same FLOPs budgets. This motivates \emph{inference scaling laws} that relate performance to inference-time compute. While these methods have been studied for reasoning-heavy tasks, their effectiveness for machine translation remains largely unexplored.

\paragraph{Quality Estimation for MT.} 
Quality estimation (QE) aims to assess the quality of a translation without reference. QE metrics such as KIWI~\cite{rei2022cometkiwi}, MetricX-QE~\cite{juraska2024metricx}, and Remedy-QE~\cite{tan2025remedy} have demonstrated strong correlations with human judgments and are widely adopted in shared tasks like WMT. As a result, these QE models are often used for data filtering~\cite{alves2024tower,tan2024uva,meng2025learn}, reranking and MBR decoding~\cite{farinhas2023empirical,rei2024tower}, reward models~\cite{xu2024contrastive,feng2025mt,tan2025remedy} in MT pipelines.

In the context of TTS, QE provides a natural mechanism for selecting the best candidate from multiple generations, as it can directly score hypotheses based on adequacy and fluency. However, using the same or similar models for both candidate selection and evaluation can lead to \emph{metric interference}~\cite{pombal2025adding}, motivating the use of diverse evaluation metrics and complementary human assessments.

\paragraph{Leveraging QE for better machine translation.}
\emph{Reranking} is a long-standing post-hoc procedure that selects a final translation from multiple candidates, using an auxiliary scoring function, such as QE models. Even though we share the same procedure, this paper differs in framing and scope: we study \emph{Test-Time Scaling}, systematically varying both the number of candidates ($N$) and model sizes to understand how inference-time compute impacts final quality, rather than optimizing a single reranking setup. 

Minimum Bayes risk (MBR) decoding~\citep{kumar2004minimum,freitag2022high,deguchi2024mbrs} instead selects the hypothesis with the highest expected utility, often using neural metrics as the utility function~\citep{freitag2022high, rei2024tower}. Quality-aware decoding (QAD) goes further by integrating quality signals directly into the generation process, e.g., steering beam search with QE scores~\citep{fernandes2022quality}. In contrast, our method isolates the effect of increasing inference-time compute with a fixed generation strategy, allowing a controlled analysis of scaling trends across models and budgets.

\section{Conclusions}
In this paper, we presented a systematic study of Test-Time Scaling (TTS) for machine translation, evaluating a simple best-of-$N$ framework across multiple model sizes, resource conditions, and compute budgets. We found that in high-resource settings, increasing $N$ for TTS consistently improves translation quality, with human evaluation confirming that small models at large $N$ can surpass larger models at $N{=}1$. These gains, however, typically require more inference compute, and under fixed TFLOP budgets larger models are generally more efficient. In low-resource settings, TTS can even degrade quality due to the limited robustness of current QE and metric models. Lastly, beyond our main findings, we show that TTS-MT is a strong yet simple baseline that can outperform recent approaches that (1) use QE models as reward signals in RLHF pipelines, or (2) employ more complex reasoning models (e.g., GPT4o-mini) for inference scaling.
We therefore encourage the community to adopt TTS-MT as a standard baseline when evaluating inference-scaling or MT quality-estimation methods.

\section*{Limitations}
We acknowledge two main limitations. First, our evaluation primarily relies on automatic MT metrics, which may introduce bias or inaccuracies compared to human judgments. Second, our experiments focus solely on the Qwen2.5 model family in zero-shot settings, and results may not generalize to other model architectures or fine-tuned systems.

Future work should also explore more robust selection methods, strategies to mitigate low-resource regressions, and adaptive approaches that jointly optimize model size and inference-time budget.

\section*{Broader Impact}
We acknowledge potential ethical considerations in MT research. To avoid the potential risks, we prioritize high-quality data from WMT24. We also acknowledge that human biases like gender and racial discrimination could exist in the MT data.

%\section*{Acknowledgments}
%This research was funded in part by the Netherlands Organization for Scientific Research (NWO) under project number VI.C.192.080. 

\bibliography{anthology,custom}
\bibliographystyle{acl_natbib}

\appendix

\section{Appendix}\label{sec:appendix}

\subsection{Generation prompts}\label{appendix:prompt}

For all experiments, we use the fixed prompt design (see Figure~\ref{fig:prompt}) considering language pair information and domain information.

\begin{figure*}[t]
\centering
\begin{minipage}{0.8\textwidth}
\label{appendix:pairwise_data}
\vspace{0.5em}
\noindent\textbf{Translation Prompt.} We use this chat template format prompt for all experiments:
\begin{lstlisting}
prompt = [
    {'role': 'user', 'content': "You are a helpful translation 
    assistant. Now translate the following {src_lang} text (in 
    {domain} domain) into natural, fluent {tgt_lang} sentence 
    while preserving the original meaning, tone, and register. 
    Please retain the lines and paragraph breaks in the transl
    -ation, and do not produce explanations or commentary in 
    your answer."},
    {'role': 'assistant', 'content': {src_sent}}
    ]
\end{lstlisting}

\noindent Where \{src\_lang\}, \{tgt\_lang\} represent source and target language, \{domain\} denotes the sentence domain, and \{src\_sent\} represent the source sentence.
\end{minipage}
\caption{Generation prompt format}
\label{fig:prompt}
\end{figure*}

\subsection{Detailed results}\label{appendix:details}

We also provide the detailed visualizations for all language pairs (en-\{zh,ja,de,es,ru\}, ja-zh) in the appendix.

\begin{figure*}[h!]
    \centering
    \includegraphics[width=\linewidth]{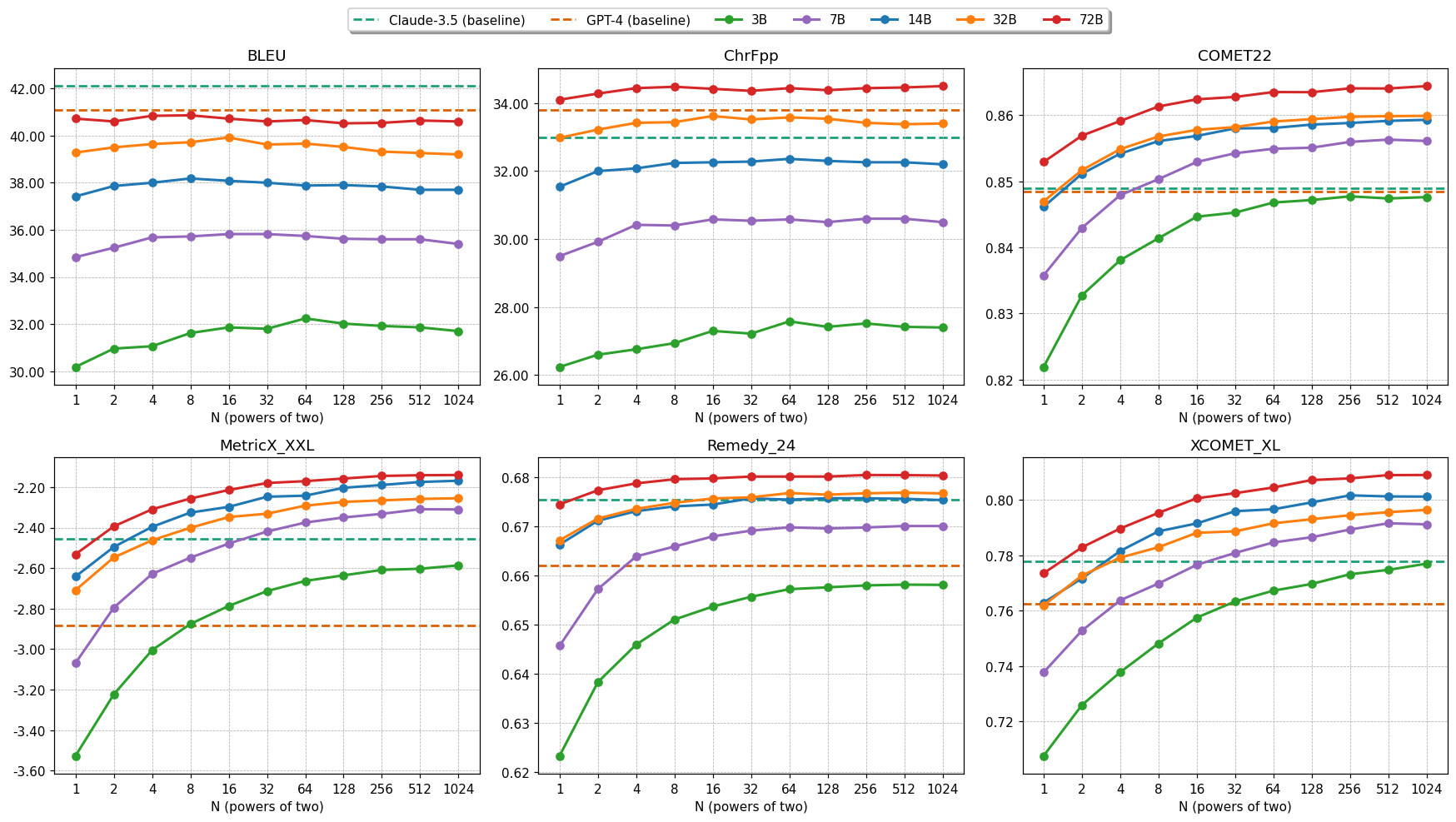}
    \caption{Test-Time-Scaling inference laws for Qwen2.5 models on English-Chinese: translation quality vs.\ number of candidates $N$ (powers of two) for best-of-$N$ across model sizes 3B-72B.}
    \label{fig:en_zh_n_grid}
    %\vspace{-2mm}
\end{figure*}

\begin{figure*}[h!]
    \centering
    \includegraphics[width=\linewidth]{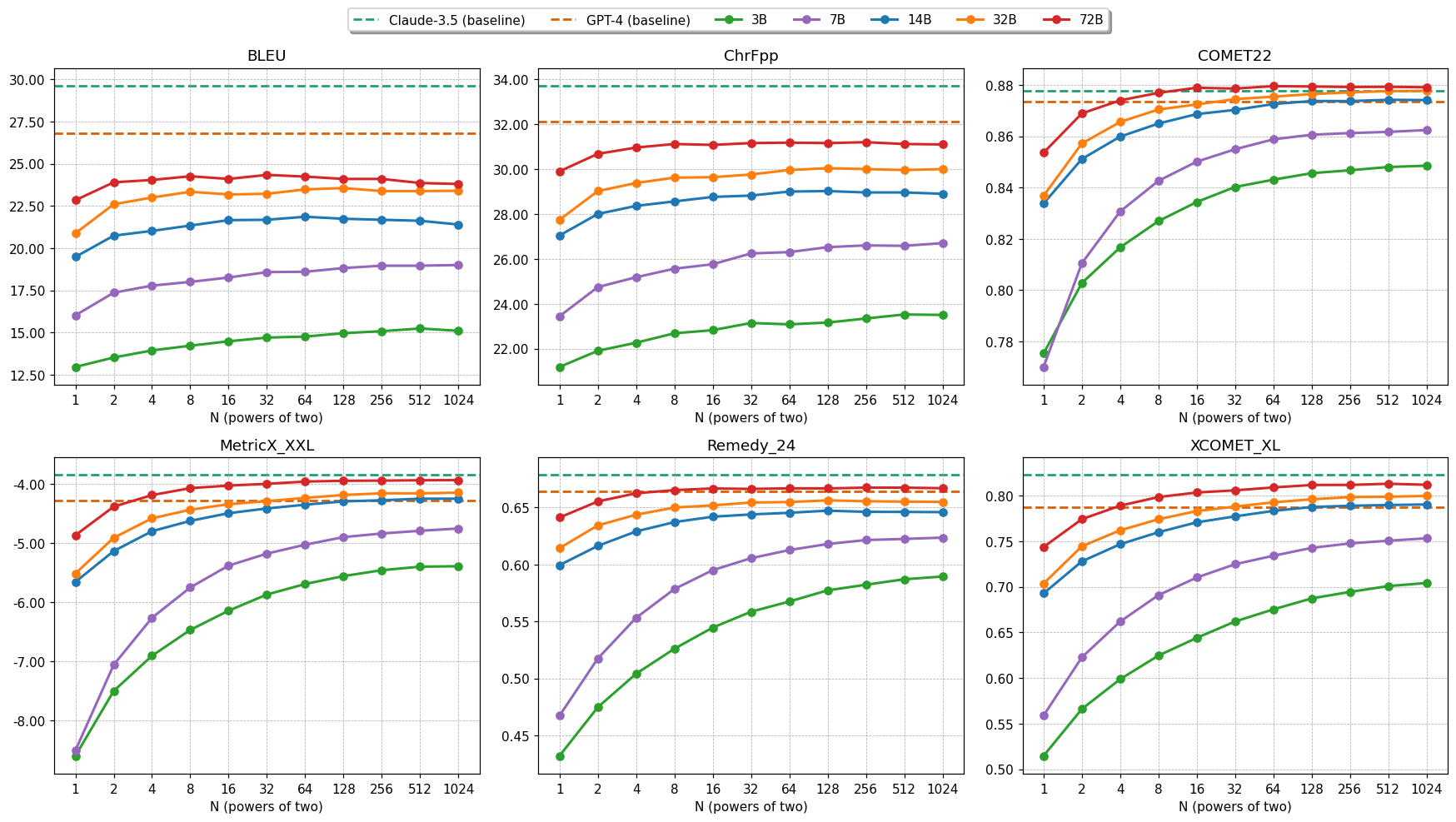}
    \caption{Test-Time-Scaling inference laws for Qwen2.5 models on English-Japanese: translation quality vs.\ number of candidates $N$ (powers of two) for best-of-$N$ across model sizes 3B-72B.}
    \label{fig:en_ja_n_grid}
    %\vspace{-2mm}
\end{figure*}

\begin{figure*}[h!]
    \centering
    \includegraphics[width=\linewidth]{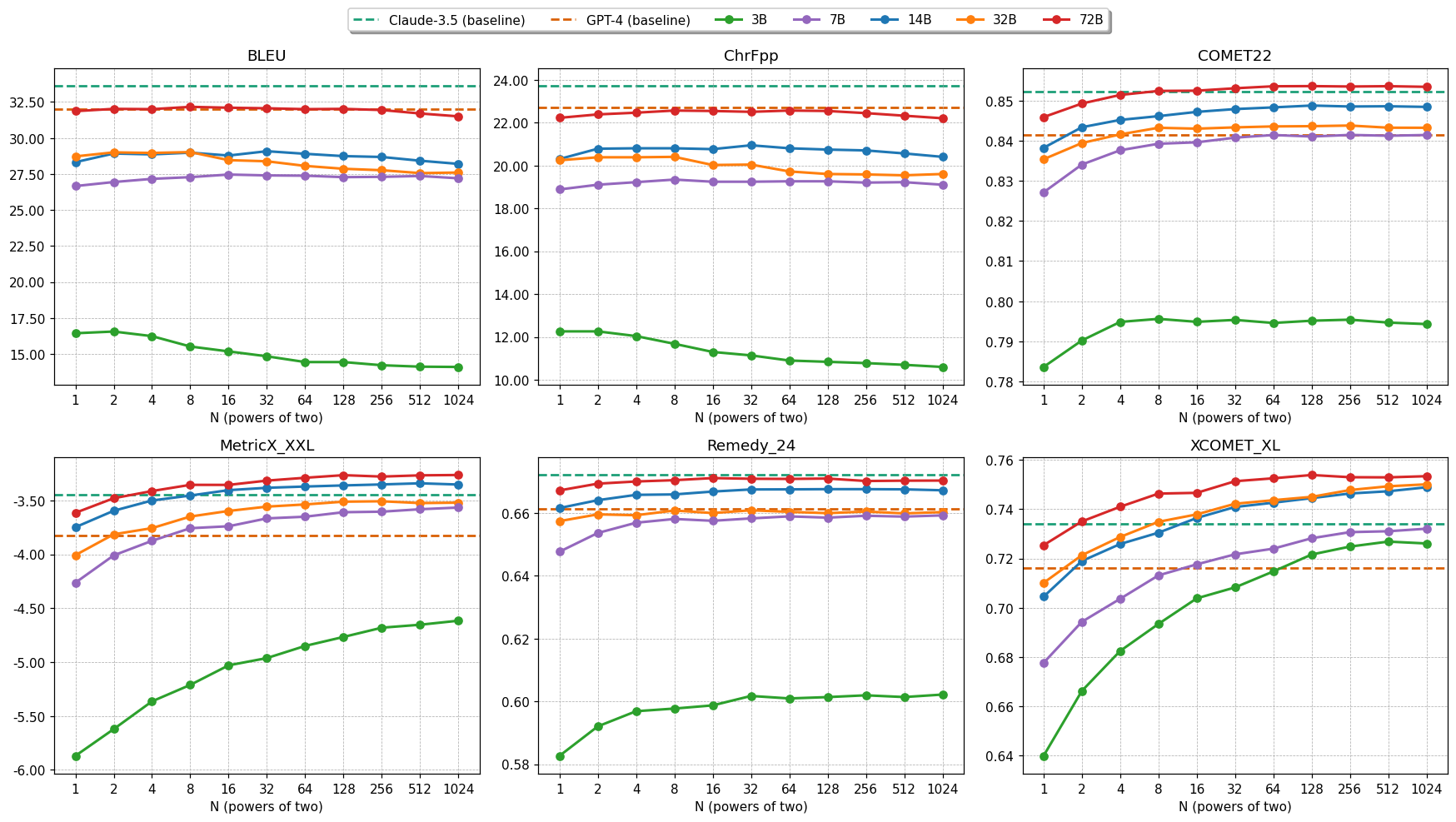}
    \caption{Test-Time-Scaling inference laws for Qwen2.5 models on Japanese-Chinese: translation quality vs.\ number of candidates $N$ (powers of two) for best-of-$N$ across model sizes 3B-72B.}
    \label{fig:ja-zh_n_grid}
    %\vspace{-2mm}
\end{figure*}

\begin{figure*}[h!]
    \centering
    \includegraphics[width=\linewidth]{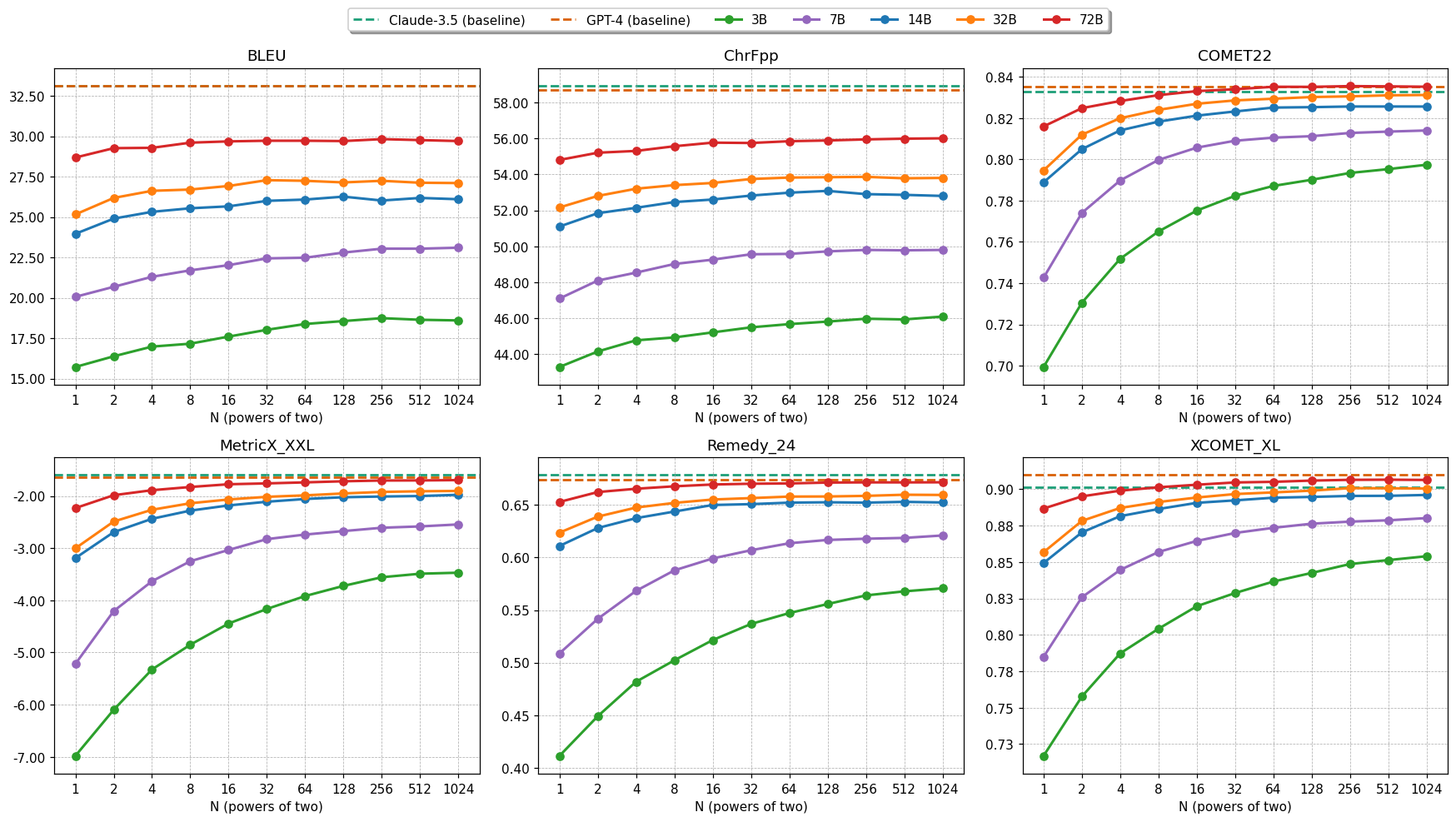}
    \caption{Test-Time-Scaling inference laws for Qwen2.5 models on English-German: translation quality vs.\ number of candidates $N$ (powers of two) for best-of-$N$ across model sizes 3B-72B.}
    \label{fig:en-de_n_grid}
    %\vspace{-2mm}
\end{figure*}

\begin{figure*}[h!]
    \centering
    \includegraphics[width=\linewidth]{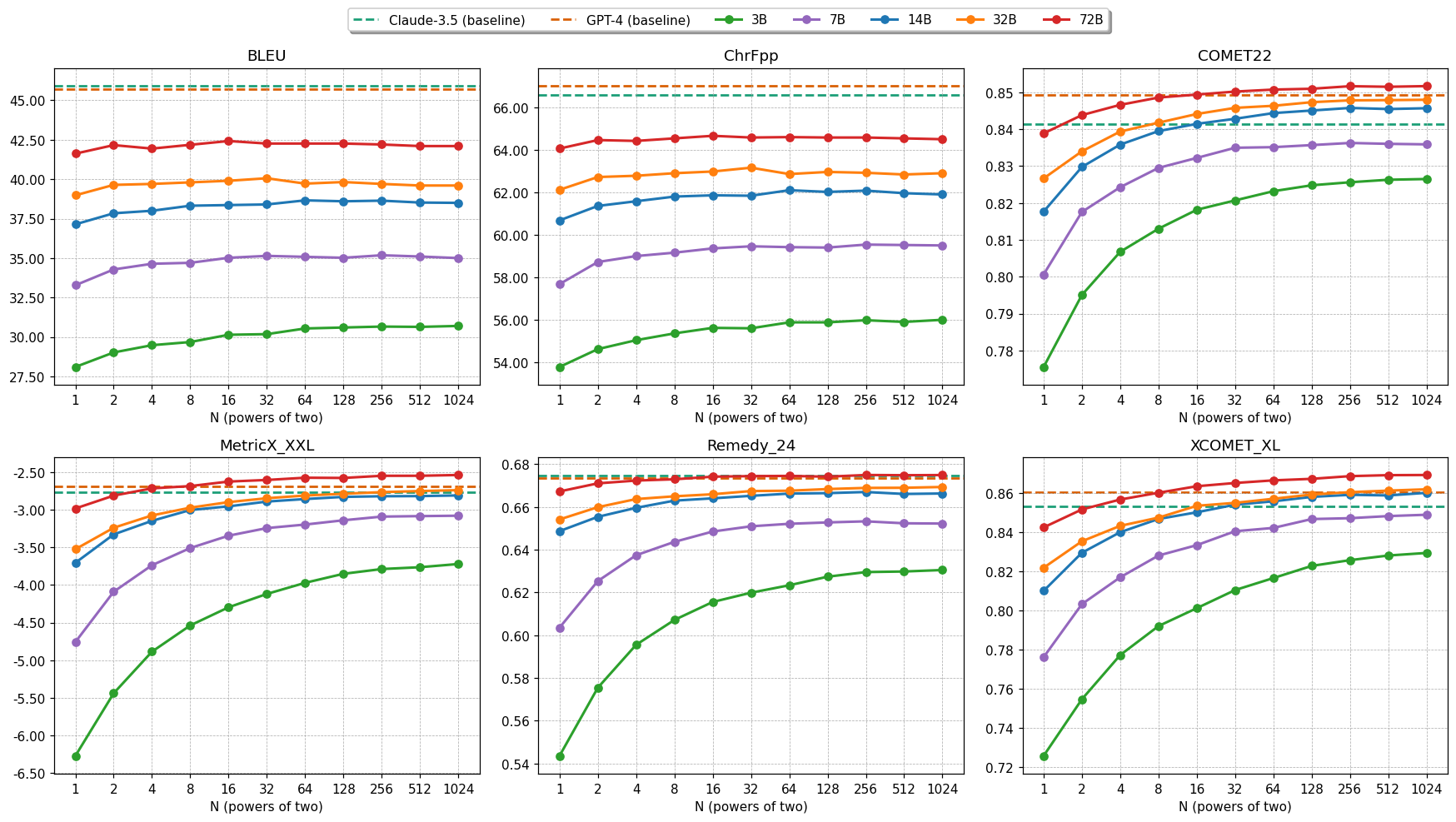}
    \caption{Test-Time-Scaling inference laws for Qwen2.5 models on English-Spanish: translation quality vs.\ number of candidates $N$ (powers of two) for best-of-$N$ across model sizes 3B-72B.}
    \label{fig:en_es_n_grid}
    %\vspace{-2mm}
\end{figure*}

\begin{figure*}[h!]
    \centering
    \includegraphics[width=\linewidth]{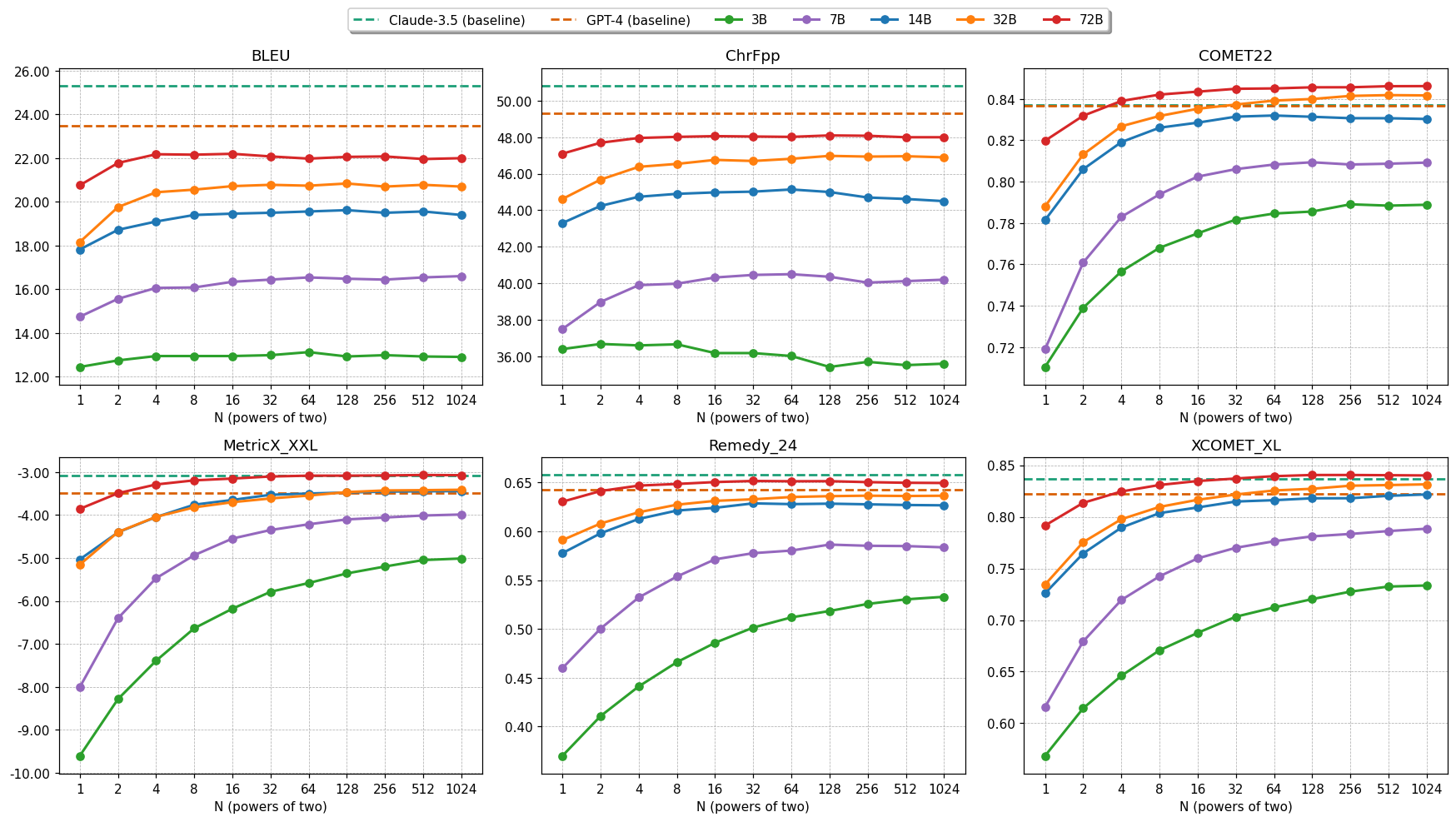}
    \caption{Test-Time-Scaling inference laws for Qwen2.5 models on English-Russian: translation quality vs.\ number of candidates $N$ (powers of two) for best-of-$N$ across model sizes 3B-72B.}
    \label{fig:en_ru_n_grid}
    %\vspace{-2mm}
\end{figure*}

\begin{figure*}[h!]
    \centering
    \includegraphics[width=\linewidth]{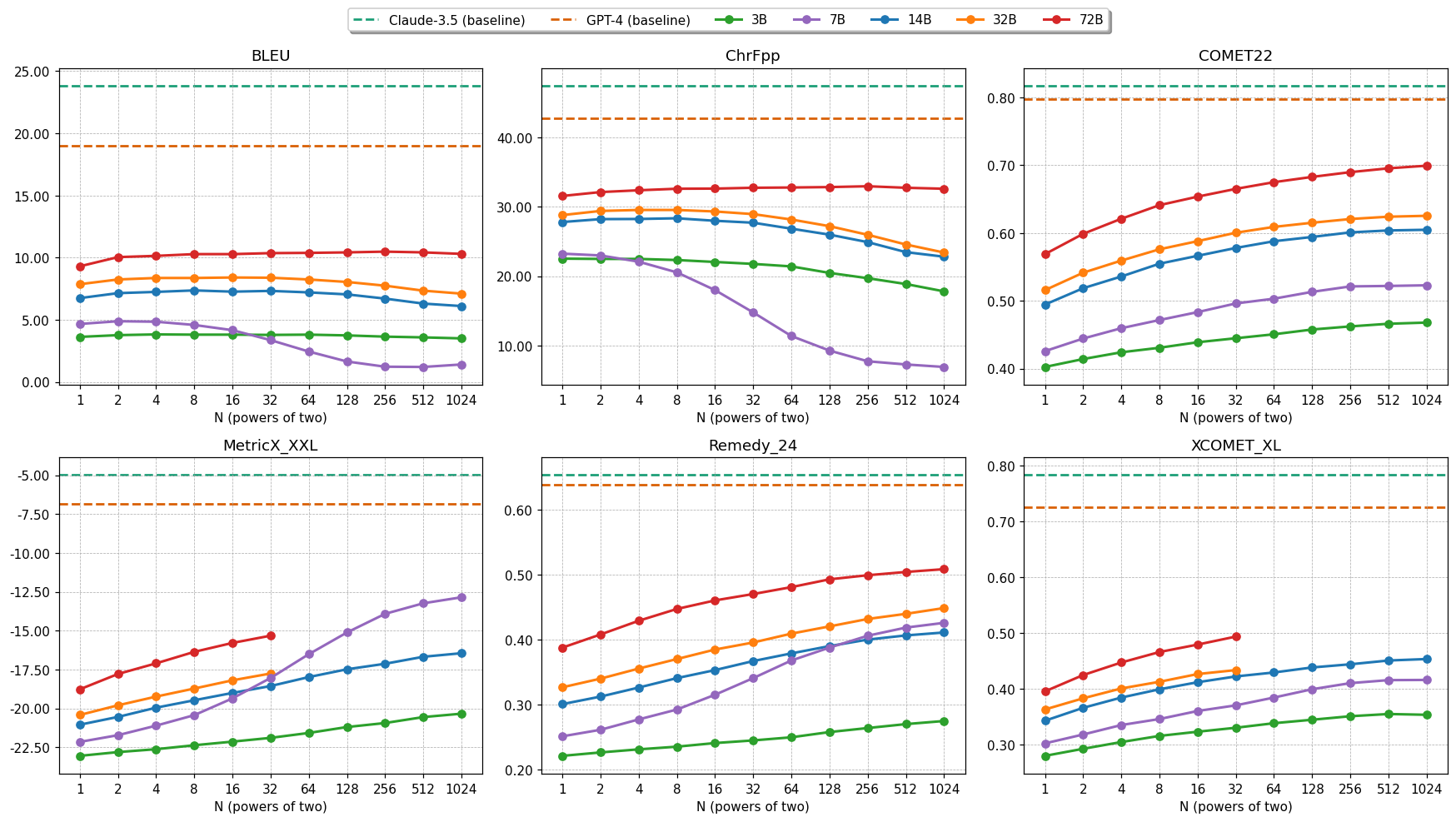}
    \caption{Test-Time-Scaling inference laws for Qwen2.5 models on English-Icelandic: translation quality vs.\ number of candidates $N$ (powers of two) for best-of-$N$ across model sizes 3B-72B.}
    \label{fig:en_is_n_grid}
    %\vspace{-2mm}
\end{figure*}

\end{document}